%% file: dog.tex
\documentclass[10pt,twocolumn,letterpaper]{article}

\usepackage{cvpr}
\usepackage{times}
\usepackage{epsfig}
\usepackage{graphicx}
\usepackage{amsmath}
\usepackage{amssymb}

\usepackage[pagebackref=true,breaklinks=true,letterpaper=true,colorlinks,bookmarks=false]{hyperref}

\usepackage{color}

\newcommand{\data}{\mbox{\sc DECADE}}

\cvprfinalcopy 

\def\cvprPaperID{603} 

\ifcvprfinal\fi
\begin{document}

\title{
Who Let The Dogs Out?\\
Modeling Dog Behavior From Visual Data\\
}

\author{Kiana Ehsani$^1$, Hessam Bagherinezhad$^1$, Joseph Redmon$^1$\\
Roozbeh Mottaghi$^2$, Ali Farhadi$^{1,2}$\\
$^1$ University of Washington, $^2$ Allen Institute for AI (AI2)}

\maketitle
\thispagestyle{empty}

\begin{abstract}
We study the task of directly modelling a visually intelligent agent. Computer vision typically focuses on solving various subtasks related to visual intelligence. We depart from this standard approach to computer vision; instead we directly model a visually intelligent agent. Our model takes visual information as input and directly predicts the actions of the agent. Toward this end we introduce \data, a dataset of ego-centric videos from a dog's perspective as well as her corresponding movements. Using this data we model how the dog \emph{acts} and how the dog \emph{plans} her movements. We show under a variety of metrics that given just visual input we can successfully model this intelligent agent in many situations. Moreover, the representation learned by our model encodes distinct information compared to representations trained on image classification, and our learned representation can generalize to other domains. In particular, we show strong results on the task of walkable surface estimation and scene classification by using this dog modelling task as representation learning. Code is available at \href{https://github.com/ehsanik/dogTorch}{https://github.com/ehsanik/dogTorch}.
\end{abstract}

\input{intro2}
\input{related}
\input{dataset}

\input{model}

\input{experiments}

\input{conclusion}

\small{\noindent\textbf{Acknowledgements:} We would like to thank Carlo C. del Mundo for his help with setting up the data collection system, Marc Milestone for his help in data collection and miss Kelp M. Redmon for being the dog in our data collection. This work is in part supported by ONR N00014-13-1-0720, NSF IIS-1338054, NSF-1652052, NRI-1637479, Allen Distinguished Investigator Award, and the Allen Institute for Artificial Intelligence. }

{\small
\bibliographystyle{ieee}
\bibliography{egbib}
}

\end{document}

%% file: intro2.tex
\section{Introduction}
 Computer vision research typically focuses on a few well defined tasks including image classification, object recognition, object detection, image segmentation, etc. These tasks have organically emerged and evolved over time as proxies for the actual problem of visual intelligence. Visual intelligence spans a wide range of problems and is hard to formally define or evaluate. As a result, the proxy tasks have served the community as the main point of focus and indicators of progress.

\begin{figure}[t]
\begin{center}
\includegraphics[width=0.95\linewidth]{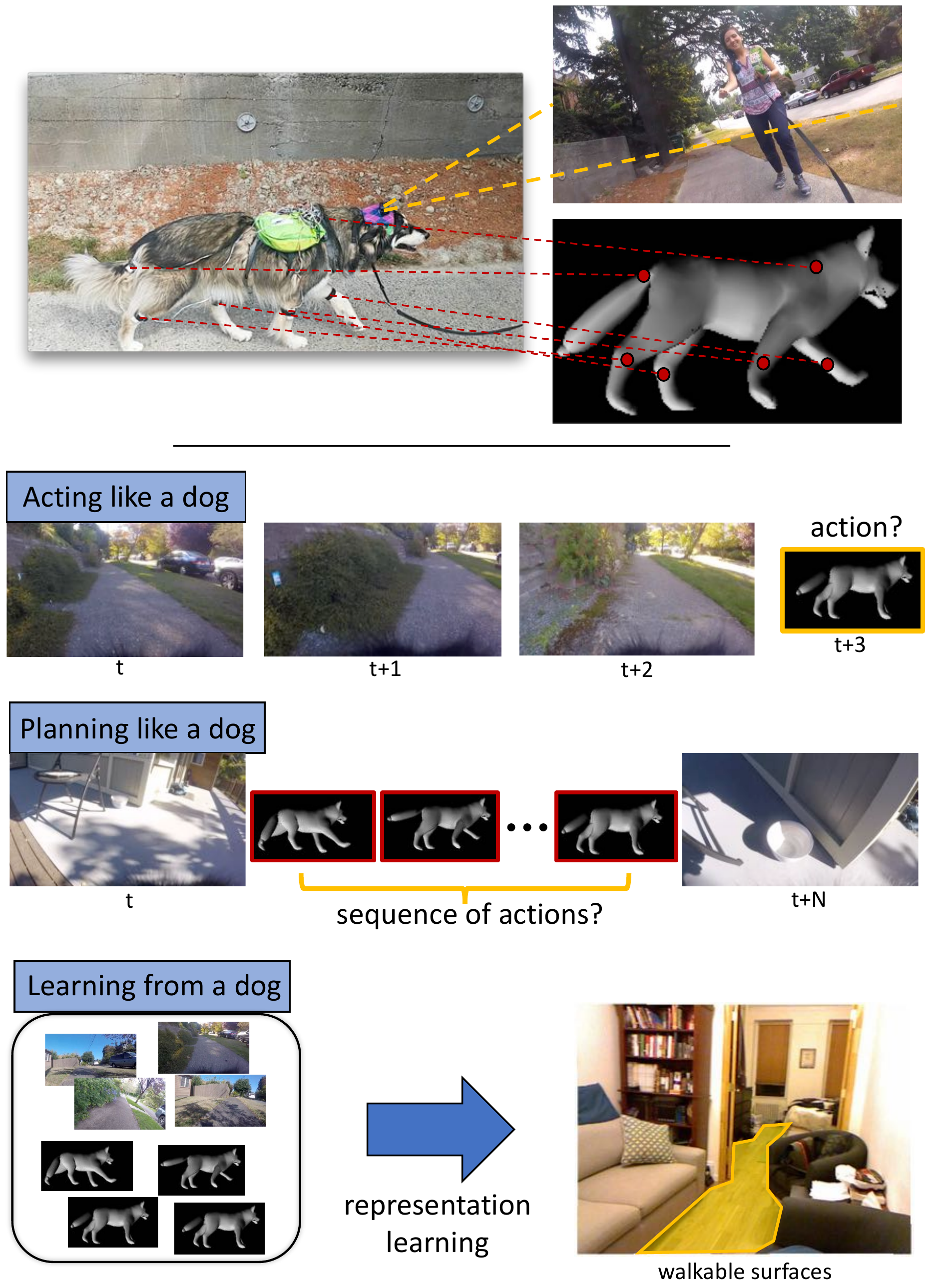}
\caption{We address three problems: (1) \emph{Acting like a dog}: where the goal is to predict the future movements of the dog given a sequence of previously seen images. (2) \emph{Planning like a dog}: where the goal is to find a sequence of actions that move the dog between the locations of the given pair of images. (3) \emph{Learning from a dog}: where we use the learned representation for a third task (e.g., walkable surface estimation). } 
\vspace{-0.7cm}
\label{fig:teaser}
\end{center}
\end{figure}

We value the undeniable impact of these proxy tasks in computer vision research and advocate the continuation of research on these fundamental problems. There is, however, a gap between the ideal outcome of these proxy tasks and the expected functionality of visually intelligent systems. In this paper, we take a direct approach to the problem of visual intelligence. Inspired by recent work that explores the role of action and interaction in visual understanding~\cite{zhu17,agrawal16,pinto16}, we define the problem of visual intelligence as \textit{understanding visual data to the extent that an agent can take actions and perform tasks in the visual world.} Under this definition, we propose to learn to act like a visually intelligent agent in the visual world.

Learning to act like visually intelligent agents, in general, is an extremely challenging and a hard-to-define problem. Actions correspond to a wide range of movements with complicated semantics. In this paper, we take a small step towards the problem of learning to directly act like intelligent agents by considering actions in their most basic and semantic-free form: simple movements.

We choose to model a dog as the visual agent. Dogs have a much simpler action space than, say, a human, making the task more tractable. However, they clearly demonstrate visual intelligence, recognizing food, obstacles, other humans and animals, and reacting to those inputs. Yet their goals and motivations are often unknown a priori. They simply exist as sovereign entities in our world. Thus we are modelling a black box where we only know the inputs and outputs of the system.

In this paper, we study the problem of learning to act and plan like a dog from visual input. We compile the Dataset of Ego-Centric Actions in a Dog Environment (\data), which includes ego-centric videos of a dog with her corresponding movements. To record movements we mount Inertial Measurement Units (IMU) on the joints and the body of the dog. We record the absolute position and can calculate the relative angle of the dog's main limbs and body.

Using \data, we explore three main problems in this paper (Figure~\ref{fig:teaser}): (1) learning to act like a dog; (2) learning to plan like a dog; and (3) using dogs movements as supervisory signal for representation learning.

In \textit{learning to act like a dog}, we study the problem of predicting the dog's future moves, in terms of all the joint movements, by observing what the dog has observed up to the current time. In \textit{learning to plan like a dog}, we address the problem of estimating a sequence of movements that take the state of the dog's world from what is observed at a given time to a desired observed state. In \textit{using dogs as supervision}, we explore the potentials of using the dogs movements for representation learning.

Our evaluations show interesting and promising results. Our models can predict how the dog moves in various scenarios (act like a dog) and how she decides to move from one state to another (plan like a dog). In addition, we show that the representation our model learns on dog behavior generalizes to other tasks. In particular, we see accuracy improvements using our dog model as pretraining for walkable surface estimation and scene recognition.

%% file: related.tex
\section{Related Work}
To the best of our knowledge there is little to no work that directly models dog behavior. We mention past work that is most relevant.

\noindent \textbf{Visual prediction.}
\cite{yuen10,pintea14} predict the motion of objects in a static image using a large collection of videos. \cite{pei11} infer the goals of people and their intended actions. \cite{ryoo11} infer future activities from a stream of video. \cite{gong11} improve tracking by considering multiple hypotheses for future plans of people. \cite{hoai12} recognize partial events, which enables early detection of events. \cite{kitani12} perform activity forecasting by integrating semantic scene understanding with optimal control theory. \cite{koppula13} use object affordances to predict the future activities of people. \cite{xie13} localize functional objects by predicting people's intent. \cite{walker14} propose an unsupervised approach to predict possible motions and appearance of objects in the future. \cite{lan14} propose a hierarchical approach to predict a set of actions that happen in the future. \cite{ranzato14} propose a method to generate the future frames of a video. \cite{kooij14} predict the future paths of pedestrians from a vehicle camera. \cite{singh2016krishnacam} predict future trajectories of a person in an ego-centric setting. \cite{mottaghi16} predict the future trajectories of objects according to Newtonian physics. \cite{vondrick16} predict visual representations for future images. \cite{zeng17} forecast future frames by learning a policy to reproduce natural video sequences. Our work is different from these works since our goal is to predict the behavior of a dog and the movement of the joints from an ego-centric camera that captures the viewpoint of the dog. 

\noindent \textbf{Sequence to sequence models.} Sequence to sequence learning \cite{sutskever14} has been used for different applications in computer vision such as representation learning \cite{srivastava15}, video captioning \cite{venugopalan15, yao15}, human pose estimation \cite{walker17}, motion prediction \cite{martinez2017human}, or body pose labeling and forecasting \cite{fragkiadaki15,walker17}. Our model fits into this paradigm since we map the frames in a video to joint movements of the dog.

\noindent \textbf{Ego-centric vision.} Our work is along the lines of ego-centric vision (e.g., \cite{fathi11,pirsiavash12,lee12,li13}) since we study the dog's behavior from the perspective of the dog. However, dogs have less complex actions compared to humans, which makes the problem more manageable. Prior work explores future prediction in the context of ego-centric vision. \cite{zhou15} infer the temporal ordering of two snippets of ego-centric videos and predict what will happen next. \cite{park16} predict plausible future trajectories of ego-motion in ego-centric stereo images.
\cite{jiang2017seeing} estimates the 3D joint position of unseen body joints using ego-centric videos.
\cite{rhinehart17} use online reinforcement learning to forecast the future goals of the person wearing the camera. In contrast, our work focuses on predicting future joint movements given a stream of video.

\noindent \textbf{Ego-motion estimation.} Our planning approach shares similarities with ego-motion learning. \cite{zhou17} propose an unsupervised approach for camera motion estimation. \cite{wang17} propose a method based on combination of CNNs and RNNs to perform ego-motion estimation for cars. \cite{melekhov17} learn a network to estimate relative pose of two cameras. \cite{ummenhofer17} also train a CNN to learn depth map and motion of the camera in two consecutive images. In contrast to these approaches that estimate translation and rotation of the camera, we predict a sequence of joint movements. Note that the joint movements are constrained by the structure of the dog body so the predictions are constrained. 

\noindent \textbf{Action inference \& Planning.} Our dog planning model infers the action sequence for the dog given a pair of images showing before and after action execution. \cite{agrawal16} also learn the mapping between actions of a robot and changes in the visual state for the task of pushing objects. \cite{pathak17} optimize for actions that capture the state changes in an exploration setting. 

\noindent \textbf{Inverse Reinforcement Learning.} Several works (e.g., \cite{abbeel04,baker09, rhinehart17}) have used Inverse Reinforcement Learning (IRL) to infer the agent's reward function from the observed behavior. IRL is not directly applicable to our problem since our action space is large and we do not have multiple training examples for each goal.

\noindent \textbf{Self-supervision.} Various research explores representation learning by different self-supervisory signals such as ego-motion \cite{agrawal15,jayaraman15}, spatial location \cite{doersch15}, tracking in video \cite{wang15}, colorization \cite{zhang16}, physical robot interaction \cite{pinto16}, inpainting \cite{pathak16}, sound \cite{owens16}, etc. As a side product, we show we learn a useful representation using embeddings of joint movements and visual signals.

%% file: dataset.tex
\begin{figure*}
\begin{center}
\includegraphics[width=0.95\linewidth]{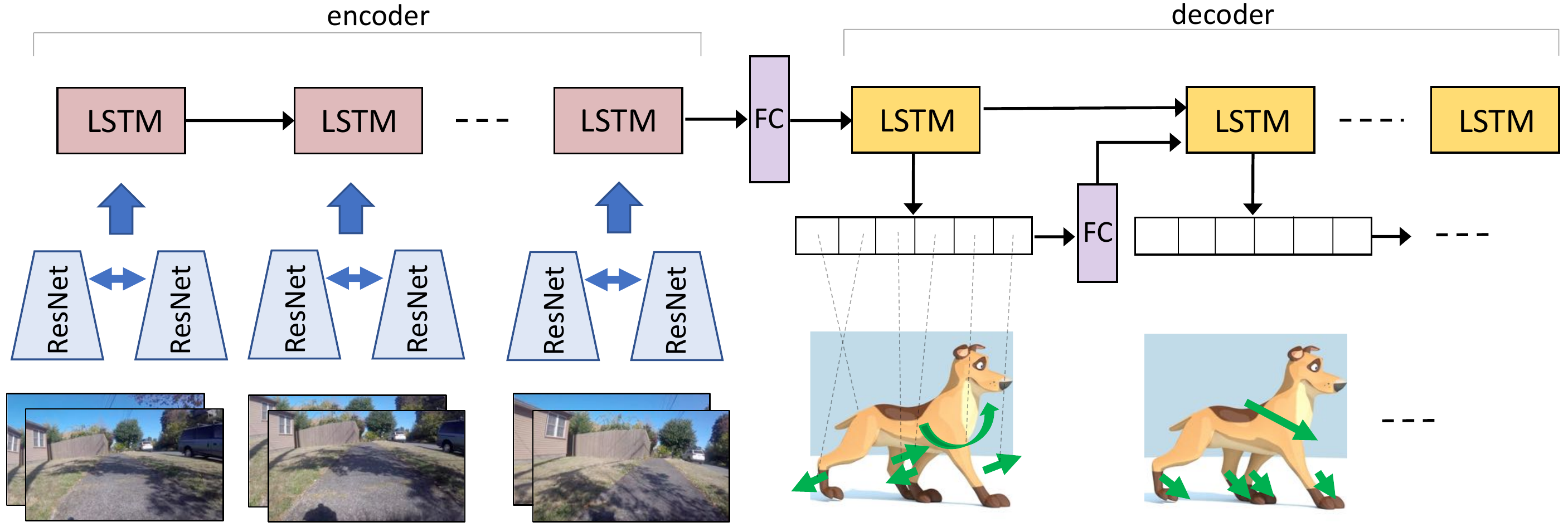}
\caption{\textbf{Model architecture for acting.} The model is an encoder-decoder style neural network. The encoder receives a stream of image pairs, and the decoder outputs future actions for each joint. There is a fully connected layer (FC) between the encoder and decoder parts to better capture the change in the domain (change from images to actions). In the decoder, the output probability of actions at each timestep is passed to the next timestep. We share the weights between the two ResNet towers.} 
\label{fig:model_acting}
\end{center}
\end{figure*}

\section{Dataset}

We introduce DECADE, a dataset of ego-centric dog video and joint movements. The dataset includes 380 video clips from a camera mounted on the dog's head. It also includes corresponding information about body position and movement. Overall we have 24500 frames. We use 21000 of them for training, 1500 for validation, and 2000 for testing.
Train, validation, and test splits consist of disjoint video clips.

We use a GoPro camera on the dog's head to capture the ego-centric videos. We sub-sample frames at the rate of 5 fps. The camera applies video stabilization to the captured stream. We use inertial measurement units (IMUs) to measure body position and movement. Four IMUs measure the position of the dog's limbs, one measures the tail, and one measures the body position. The IMUs enable us to capture the movements in terms of angular displacements.

For each frame, we have the absolute angular displacement of the six IMUs. Each angular displacement is represented as a $4$ dimensional quaternion vector. More details about angular calculations in this domain and the method for quantizing the data is explained in detail in Section~\ref{sec:experiment}. The absolute angular displacements of the IMUs depend on what direction the dog is facing. For that reason, we compute the difference between angular displacements of the joints, also in the quaternion space. The difference of the angular displacements between two consecutive frames (that is $0.2s$ in time) represents the action of the dog in that timestep.

An Arduino on the dog's back connects to the IMUs and records the positional information. It also collects audio data via a microphone mounted on the dog's back. We synchronize the GoPro with the IMU measurements using audio information. This allows us to synchronize the video stream with the IMU readings with microsecond precision. 
We collect the data in various outdoor and indoor scenes: living room, stairs, balcony, street, and dog park are examples of these scenes. The data is recorded in more than 50 different locations. We recorded the behavior of the dog while involved in certain activities such as walking, following, fetching, interaction with other dogs, and tracking objects. No annotations are provided for the video frames, we use the raw data for our experiments.

%% file: model.tex
\section{Acting like a dog}
\label{sec:act}
We predict how the dog acts in the visual world in response to various situations. Specifically, we model the future actions of the dog given a sequence of previously seen images.
The input is a sequence of image frames $(I_1, I_2,\ldots, I_t)$, and the output is the future actions (movements) of each joint $j$ at each timestep $t < t' \le N$: $(a^j_{t+1}, a^j_{t+2}, \ldots, a^j_{t+N})$. Timesteps are spaced evenly by $0.2s$ in time. The action $a^j_t$ is the movement of the joint $j$, that along with the movements of other joints, takes us from image frame $I_t$ to $I_{t+1}$. For instance, $a^2_{3}$ represents the movement of the second joint that takes place between image frames $I_{3}$ and $I_{4}$. Each action is the change in the orientation of the joints in the 3D space.

We formulate the problem as classification, i.e. we quantize joint angular movements and label each joint movement as a ground-truth action class. To obtain action classes, we cluster changes in IMU readings (joint angular movements) by $K$-means, and we use quaternion angular distances to represent angular distances between quaternions. Each cluster centroid represents a possible movement of that joint.

Our movement prediction model is based on an encoder-decoder architecture, where the goal is to find a mapping between input images and future actions. For instance, if the dog sees her owner with a bag of treats, there is a high probability that the dog will sit and wait for a treat, or if the dog sees her owner throwing a ball, the dog will likely track the ball and run toward it.

Figure~\ref{fig:model_acting} shows our model. The encoder part of the model consists of a CNN and an LSTM. At each timestep, the CNN receives a pair of consecutive images as input and provides an embedding, which is used as the input to the LSTM. That is, the LSTM cell receives the features from frames $t$ and $t+1$ as the input in a timestep, and receives frames $t+1$ and $t+2$ in the next timestep. Our experimental results show that observing the two frames in each timestep of LSTM improves the performance of the model. The CNN consists of two towers of ResNet-18~\cite{resnet}, one for each frame, whose weights are shared.

The decoder's goal is to predict the future joint movements of the dog given the embedding of the input frames. The decoder receives its initial hidden state and cell from the encoder. At each timestep, the decoder outputs the action class for each of the joints. The input to the decoder at the first timestep is all zeros, at all other timesteps, we feed in the prediction of the last timestep, embedded by a linear transformer. Since we train the model with fixed output length, no stop token is required and we always stop at a fixed number of steps. Note that there are a total of six joints; hence our model outputs six classes of actions at each timestep.

Each image is given to the ResNet tower individually and the features for the two images are concatenated. The combined features are embedded into a smaller space by a linear transformation. The embedded features are fed into the encoder LSTM. We use a ResNet pre-trained on ImageNet~\cite{imagenet} and we fine-tune it under a Siamese setting to estimate the joints movements between two consecutive frames. We use the fine-tuned ResNet in our encoder-decoder model.

We use an average of weighted class entropy losses, one for each joint, to train our encoder-decoder. Our loss function can be formulated as follows:
\begin{equation}
\label{eq:loss}
    L(o, g) = \frac{1}{NK}\sum_{t=1}^N\sum_{i=1}^K \frac{1}{f^i_{g_i}}\log o(t)^i_{g(t)_i},
\end{equation}
where $g(t)_i$ is the ground-truth class for $i$-th joint at timestep $t$, $o(t)^i_{g_i}$ is the predicted probability score for $g_i$-th class of $i$-th joint at timestep $t$, $f^i_{g_i}$ is the number of data points whose $i$-th joint is labeled with $g_i$, $K$ is the number of joints, and $N$ is the number of timesteps. The $\frac{1}{f^i_{g_i}}$ factor helps the ground-truth labels that are underrepresented in the training data. 

\section{Planning like a dog}
\label{sec:plan}
Another goal is to model how dogs plan actions to accomplish a task.
To achieve this, we design a task as follows: Given a pair of non-consecutive image frames, plan a sequence of joint movements that the dog would take to get from the first frame (starting state) to the second frame (ending state). Note that a traditional motion estimator would not work here. Motion estimators infer a translation and rotation for the camera that can take us from an image to another; in contrast, here we expect the model to plan for the actuator, with its set of feasible actions, to traverse from one state to another.

More formally, the task can be defined as follows. Given a pair of images $(I_1, I_N)$, output an action sequence of length $N-1$ for each joint, that results in the movement of the dog from the starting point, where $I_1$ is observed, to the end point, where $I_N$ is observed.

\begin{figure}[t]
\begin{center}
\includegraphics[width=1.0\linewidth]{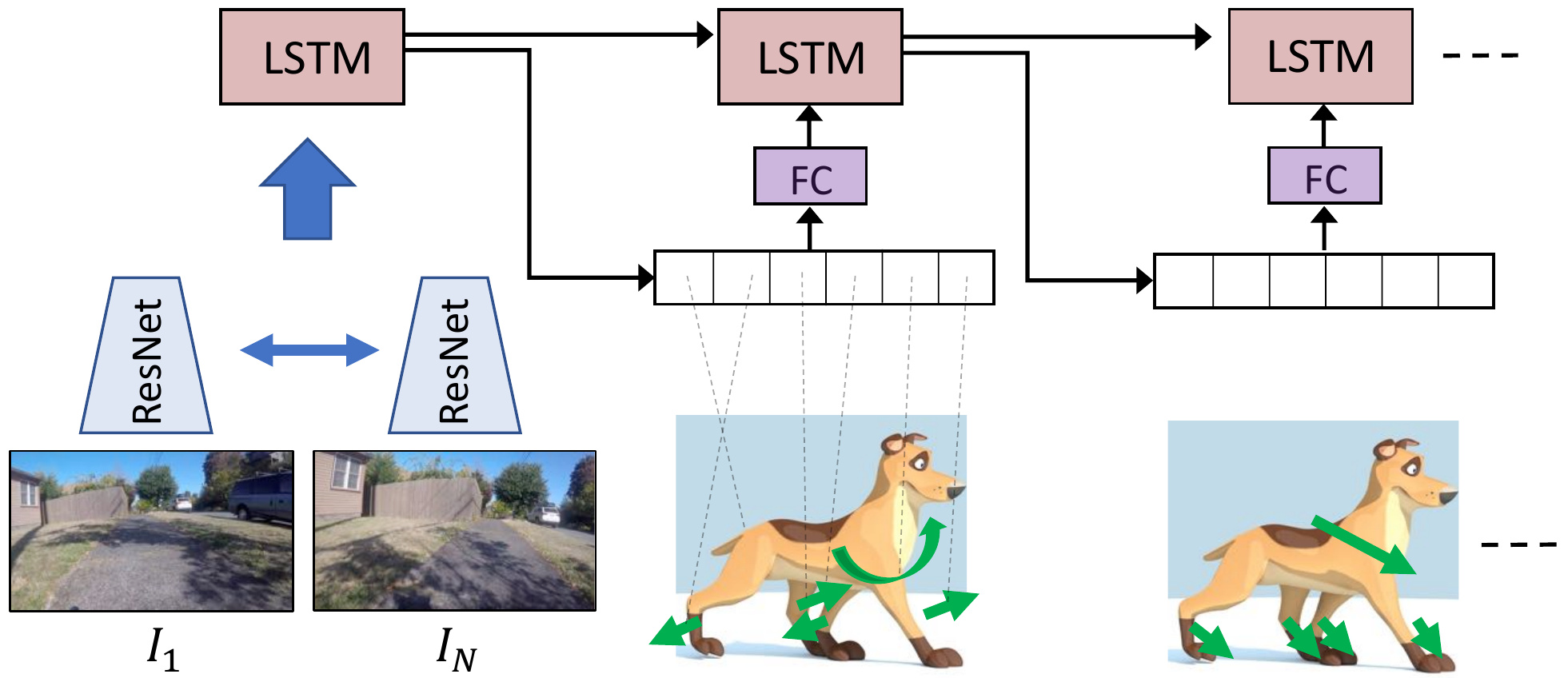}
\caption{\textbf{Model architecture for planning.} The model is a combination of CNNs and an LSTM. The inputs to the model are two images $I_1$ and $I_N$, which are $N-1$ time steps apart in the video sequence. The LSTM receives the features from the CNNs and outputs a sequence of actions (joint movements) that move the dog from $I_1$ to $I_N$. } 
\vspace{-0.3cm}
\label{fig:model_planning}
\end{center}
\end{figure}

Each action that the dog takes changes the states of the world, and therefore planning for the next steps. Thus, we design a recurrent neural network, containing an LSTM that observes the actions taken by the model in previous timesteps for the next timestamp action prediction. Figure~\ref{fig:model_planning} shows the overview of our model. We feed-forward image frames $I_1$ and $I_N$ to individual ResNet-18 towers, concatenate the features from the last layer and feed it to the LSTM. At each timestep, the LSTM cell outputs planned actions for all six joints. We pass the planned actions for a timestep as the input of the next timestep.
This enables the network to plan the next movements conditioned on the previous actions. As opposed to making hard decisions about the previously taken actions, we pass the action probabilities as the input to the LSTM in the next timestep. A low probability action at the current timestep might result in a high probability trajectory further along in the sequence. Using action probabilities prevents early pruning to keep all possibilities for the future actions.

We train this recurrent neural network using a weighted cross entropy loss over all time steps and joints as described in Equation~\ref{eq:loss}. 
Similar to the \emph{acting} problem, we use a discretized action space, which is obtained using the procedure described in Section~\ref{sec:experiment}.

\section{Learning from a dog.} 
\begin{figure}[t]
\begin{center}
\includegraphics[width=0.5\linewidth]{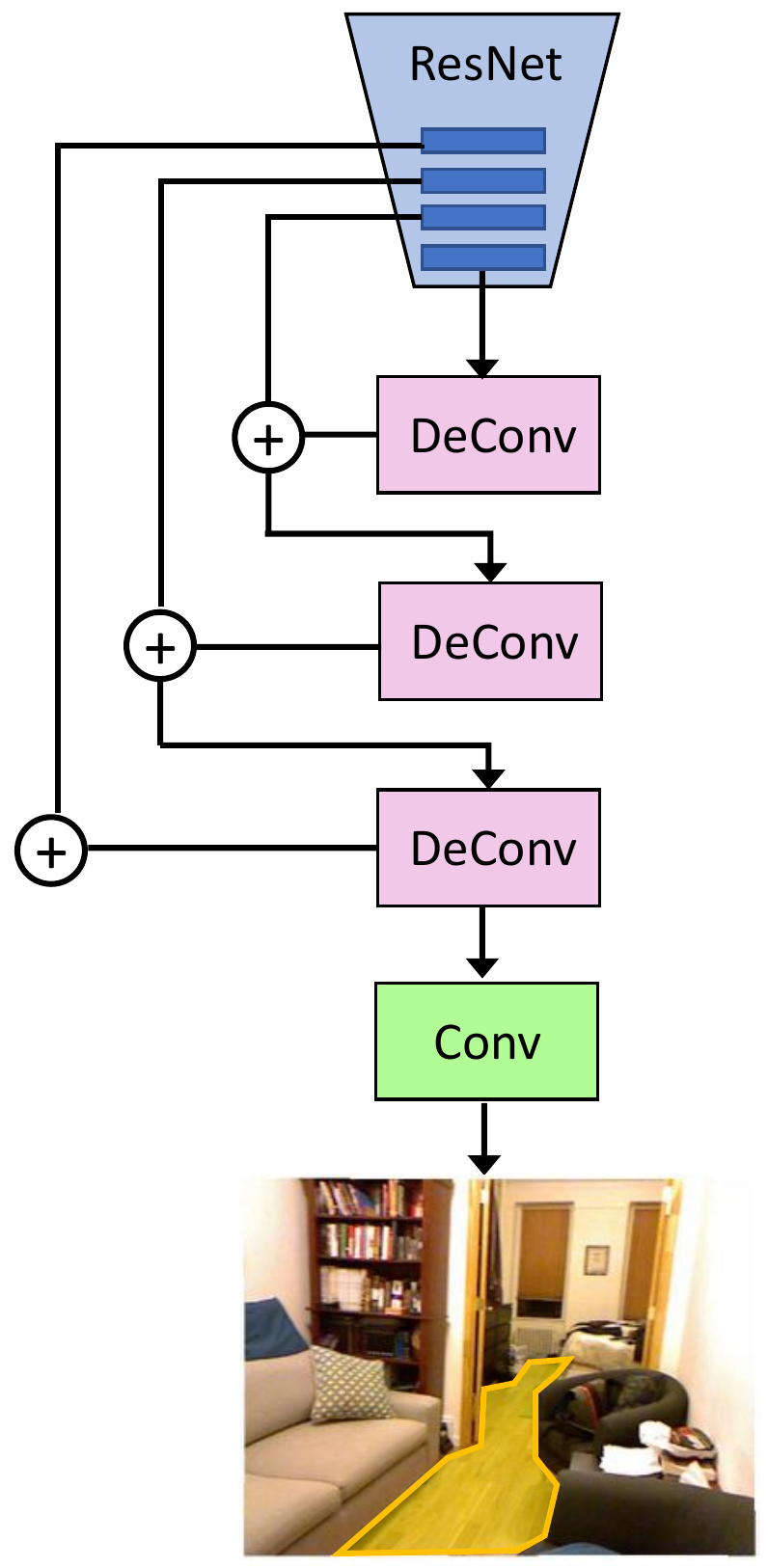}
\caption{\textbf{Model architecture for walkable surface estimation.} We augment the last four layers of ResNet with Deconvolution and Convolution layers to infer walkable surfaces. } 
\vspace{-0.3cm}
\label{fig:walkable}
\end{center}
\end{figure}

While learning to predict the movements of the dog's joints from the images that the dog observes we obtain an image representation that encodes different types of information. To learn a representation, we train a ResNet-18 model to estimate the current dog movements (the change in the IMUs from time $t-1$ to $t$) by looking at the images that the dog observes in time $t-1$ and $t$. 
We then test this representation, and compare with a ResNet-18 model trained on ImageNet, in a different task using separate data. For our experiments we choose the task of walkable surface estimation~\cite{mottaghi16b} and scene categorization using SUN397 dataset \cite{sun397}. Figure~\ref{fig:walkable} depicts our model for estimating the walkable surfaces from an image. To showcase the effects of our representation, we replace the ResNet-18 part of the model shown in blue with a ResNet trained on ImageNet and compare it with a ResNet trained on \data. 

%% file: experiments.tex
\section{Experiments}
\label{sec:experiment}

We evaluate our models on (1) how well they can predict the future movements of the dog (acting), (2) how accurately they can plan like a dog, and (3) how well the representation learned from dog data generalizes to other tasks.

\subsection{Implementation details}

We use inertial measurement units (IMUs) to obtain the angular displacements of the dog's joints. The IMUs that we use in this project process the angular displacements internally and provide the absolute orientation in quaternion form at average rate of 20 readings per second. To synchronize all IMUs together, we connect all the IMUs to the same embedded system (Raspberry pi 3.0). We use a GoPro on the dog's head to capture ego-centric video, and we sub-sample the images at a rate of 5 frames per second. To sync the GoPro and Raspberry pi, we use audio that has been recorded on both instruments.

The rate of the joint movement readings and video frames are different. We perform interpolation and averaging to compute the absolute angular orientation for each frame. For each frame of the video, we compute the average of IMU readings, in quaternion space, corresponding to a window of 0.1 second centered at the current frame.

To factor out the effects of global orientation change we use the relative orientations rather than the absolute ones. We compute the difference of absolute angular orientations corresponding to consecutive frames after calculating the average of quaternions for each frame. The reason for using quaternions, instead of Euler angles, is that subtracting two quaternions is more well-defined and is easily obtained by:
\[
q_2 - q_1 = q_1^{-1}q_2
\]

We use K-means clustering to quantize the action space (space of relative joint movements). The distance function that we use for K-means clustering is defined by:
\[
\text{dist}(q_1, q_2) = 2 \text{arccos}(\langle q_1, q_2\rangle)
\]

The reason for formulating the problem as classification rather than regression is that our experimental evaluations showed that CNNs obtain better results for classification (as opposed to regression to continuous values). The same observation has been made by \cite{of_iccv2015, Wang_deep3d_2015, mottaghi16_whi}, where they also formulate a continuous value estimation problem as classification. 

We treat each joint separately during training. It is more natural to consider all joints together for classification to respect the kinematic constraints, however it should be noted that: (1) It is hard to collect enough training data for all possible joint combinations (3910 different joint configurations appear in our training data); (2) By combining the losses for all joints, the model encodes an implicit model of kinematic possibilities; (3) Our experiments showed that per-joint clustering is better than all-joint clustering. 

To visualize the dog movements we use a 3D model of a dog from \cite{zuffi17}. Their model is a 3D articulated model that can represent animals, such as dogs. For visualization, we apply the movement estimates from the model to the dog model.

\noindent\textbf{Learning to Act.} The input to the acting network, explained in Section~\ref{sec:act}, are pairs of frames of size $224\times224$ and the output is a sequence of movements predicted for future actions of the dog. The input images are fed into two ResNet-18 towers with shared weights. The outputs of the ResNets (the layer before the classification layer), which are of size $512$, are concatenated into a vector of size $1024$. The image vector is then used as the input to the encoder LSTM. The encoder LSTM has a hidden size of $512$, and we set the initial hidden and cell states to all zeros. 

The hidden and cell state of the last LSTM cell of the encoder are used as the initialization of the hidden and cell state of the LSTM for the decoder part. There is a fully connected layer before the input to the decoder LSTM to capture the domain change between the encoder and the decoder (encoder is in the image domain, while the decoder in the joint movement domain).

\begin{table}[t]
\centering
\begin{tabular}{|l | c | }
 \hline
 Model &    Test Accuracy \\
 \hline
 Nearest Neighbor &  $13.14$ \\
 \hline
 CNN - regression &  $12.73$\\
 \hline
 Our Model -- Single Tower &   $18.65$\\
 \hline
 Our Model &   $\mathbf{20.69}$\\
\hline
\end{tabular} 
\vspace{0.1cm}
\caption{Inferring the action between two consecutive frames.}
\vspace{-0.2cm}
\label{tab:estimation}
\end{table}

The output of each decoder LSTM cell is then fed into 6 fully connected layers, where each one estimates the action class for each joint. We consider 8 classes of actions for each joint.  We visualized our training data for different number of clusters and observed that 8 clusters provide a reasonable separation of clusters, does not result in false clusters, and clusters correspond to natural movements of the limbs.

We pre-train the ResNets by fine-tuning them for the task of estimating the joint actions, where we use a pair of consecutive frames as input and we have 6 different classification layers corresponding to different joints.

\noindent\textbf{Learning to Plan.} For the planning network, the input is obtained by concatenation of the ResNet-18 features for the source and destination input images (a vector of size $2048$). A fully connected layer receives this vector as input and converts it to a 512 dimensional vector, which is then used as the first time step input for the LSTM. The LSTM output is 48 dimensional (6 joints $\times$ 8 action class). The output is followed by a $48\times512$ fully connected layer. The output of the fully connected layer is used as the input of the LSTM at the next timestep.

\noindent\textbf{Learning from a dog.}  To obtain the representation, we train a ResNet-18 model to estimate the dog movements from time $t-1$ to time $t$ by looking at the images at time $t-1$ and $t$. We use a simple Siamese network with two ResNet-18 towers whose weights are shared. We concatenate the features of both frames into a $1024$-dimensional vector and use a fully connected layer to predict the final $48$ labels ($6$ IMUs each having $8$ class of values). Table~\ref{tab:estimation} shows our results on how well this network can predict the current (not future) movements of the dog. The evaluation metric is the class accuracy described below. We use this base network to obtain our image representation. We also use this network for initializing our acting and planning networks.

\subsection{Evaluation metrics}
We use different evaluation metrics to compare our method with the baselines. 

\noindent\textbf{Class Accuracy:} This is the standard metric for classification tasks. We report the average per class accuracy rather than the overall accuracy for two reasons: 1) The dataset is not uniformly distributed among the clusters and we do not want to favor larger clusters over smaller ones, and 2) Unlike the overall unbalanced accuracy, the mean class accuracy of a model that always predicts the mode class is not higher than chance. 

\noindent\textbf{Perplexity:} Perplexity measures the likelihood of the ground-truth label. Perplexity is commonly used for sequence modeling. We report perplexity for all of the baselines and models that are probabilistic and predict a sequence. If our model assigns probability $p$ to a sequence of length $n$, the perplexity measure is calculated as $p^\frac{1}{n}$.

\begin{table}[t]
\centering
\setlength{\tabcolsep}{3pt}
\begin{tabular}{|l | c | c |}
 \hline
 Model &    Test Accuracy       &   Perplexity\\
 \hline
 Nearest Neighbor   &  $12.64$  & \texttt{N/A}\\
 \hline
 CNN   &  $19.84$   & $0.2171$\\
 \hline
 Our Model--1 tower & $18.04$ &   $0.2023$\\
 \hline
 Our Model--1 frame/timestep & $19.28$ & $0.242$\\
 \hline
 Our Model &    $\mathbf{21.62}$ & $\mathbf{0.2514}$\\
 \hline
\end{tabular}
\vspace{0.1cm}
\caption{\textbf{Acting Results.} We observe a video sequence of 5 frames and predict the next 5 actions.}
\vspace{-0.3cm}
\label{tab:prediction}
\end{table}

\begin{figure*}
\begin{center}
\includegraphics[width=0.97\linewidth]{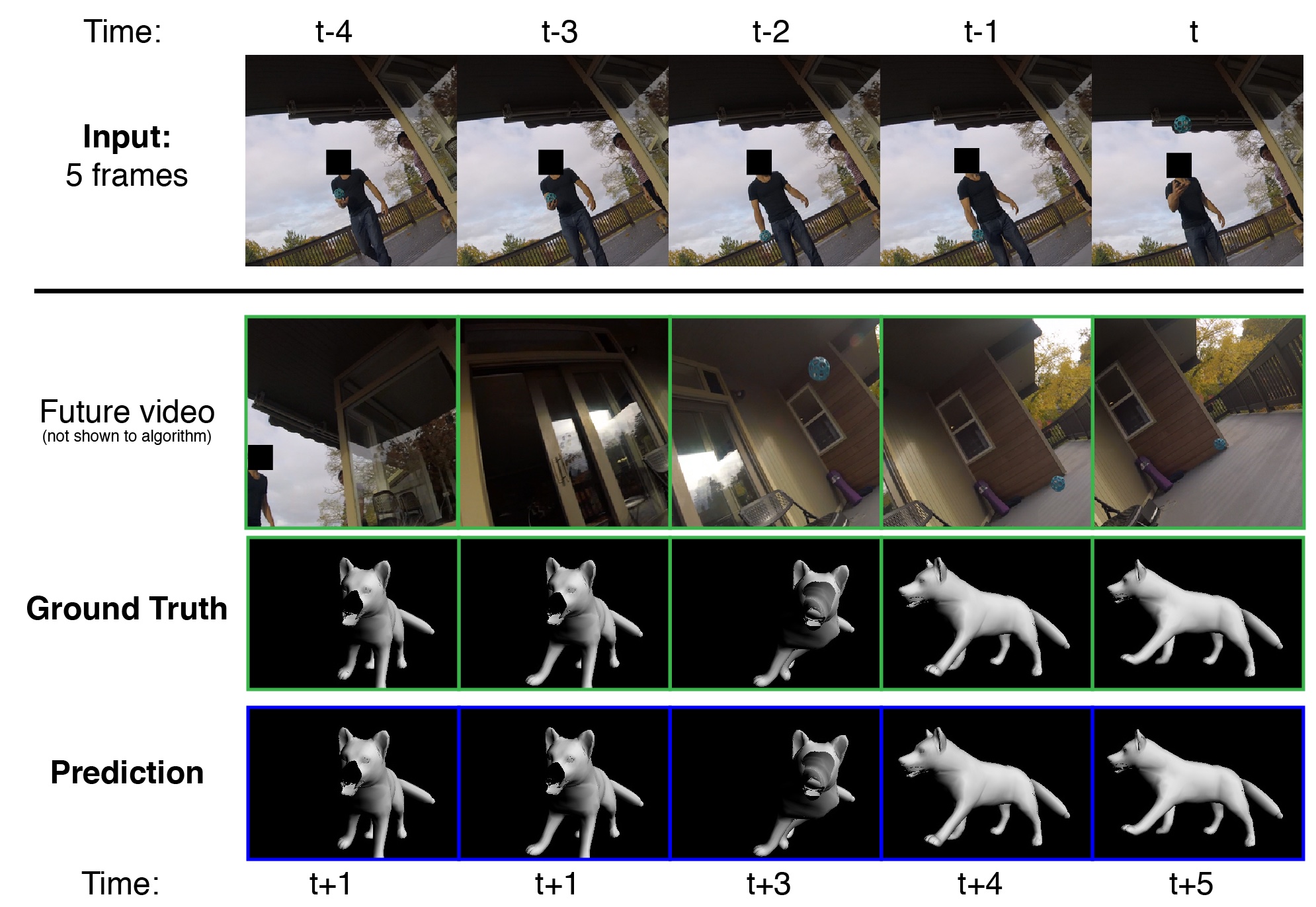}
\caption{\textbf{Qualitative Results: Learning to Act.} Our model sees 5 frames of a video where a man begins throwing a ball past the dog. In the video the ball bounces past the dog and the dog turns to the right to chase the ball. Using just the first 5 frames our model correctly predicts how the dog turns to the right as the ball flies past.} \vspace{-0.5cm}
\label{fig:results_acting}
\end{center}
\end{figure*}

\subsection{Results}
\noindent\textbf{Learning to act like a dog.} Table~\ref{tab:prediction} summarizes our experimental results for predicting the future movements of the dog using the images the dog observes. We compare our method with two baselines in terms of both test set accuracy and perplexity. The Nearest Neighbors baseline uses the features obtained from a ResNet18 network trained on ImageNet. The CNN baseline concatenates all the images into an input tensor for a ResNet18 model that classifies the future actions. We also report two ablations of our model. The 1 tower ablation uses the model depicted in Figure~\ref{fig:model_acting} but only uses one ResNet-18 tower with both frames concatenated ($6$ channels) as the input. We also compare our model with an ablation that only uses one image at each timestep instead of looking at two. Our results show that our model outperforms the baselines. Our ablations also show the importance of different components in our model. Figure~\ref{fig:results_acting} shows an example where a man is throwing a ball past the dog. Our model correctly predicts the future dog movements (the bottom row) by only observing the images the dog observed in the previous time steps (the first row). The second row shows the set of images the dog actually observed. These images were not shown to the algorithm and are depicted here to better render the situation. 

\begin{table}[t]
\centering
\begin{tabular}{|l | c | c |}
 \hline
 Model &    Test Accuracy   & Perplexity\\
 \hline
 Nearest Neighbor   &  $14.09$ & \texttt{N/A}\\
 \hline
 CNN   &  $14.61$   &  $0.1419$\\
 \hline
 Our Model &  $\mathbf{19.77}$    &   $\mathbf{0.2362}$\\
 \hline
\end{tabular}
\vspace{0.1cm}
\caption{\textbf{Planning Results.} Planning between the start and end frame. We consider start and end images that are 5 steps apart.}
\vspace{-0.3cm}
\label{tab:planning}
\end{table}

\noindent\textbf{Learning to plan like a dog.} 
Table~\ref{tab:planning} shows our experimental results for the task of planning. The nearest neighbor baseline concatenates the image features obtained from a ResNet-18 trained on ImageNet and searches for a plan of the required size that corresponds to the closest feature. The CNN baseline concatenates the input source and destination images into an input tensor that feeds into a ResNet-18 that predicts a plan for the required size. Our results show that our model outperforms these baselines in the challenging task of planning like a dog both in terms of accuracy and perplexity. 

\begin{table}[t]
\setlength{\tabcolsep}{3pt} 
\centering
\begin{tabular}{|l|c||c|}
\hline
Model & Angular metric & All joints \\ \hline
Random & 131.70  & 4e-4  \\ 
\hline
CNN-acting & 63.42 &	8.67\\ 
\hline
Our model-acting & \textbf{59.61} &	\textbf{9.49} \\
\hline
\hline
CNN-planning & 76.18 & 0.14  \\ \hline
Our model-planning & \textbf{63.14} & \textbf{3.66} \\ \hline 
\end{tabular}
\vspace{0.1cm}
\caption{Continuous evaluation and all-joint evaluation. Lower is better in the first column. Higher is better in the second column.}
\vspace{-0.3cm}
\label{tab:new-metric}
\end{table}

To better understand the behavior of the model, for both acting and planning, we show the performance in terms of a continuous angular metric and also for all joints in Table~\ref{tab:new-metric}. The angular metric compares the mean of the predicted cluster with the actual continuous joint movements in groundtruth ($\arccos (2 (q_{pred}.q_{gt}) ^ 2 - 1)$), where $q_{pred}$ and $q_{gt}$ are the predicted and groundtruth quaternion angles, respectively. The all-joint metric calculates the percentage of correct predictions, where we consider a prediction correct if all joints are predicted (classified) correctly.

\noindent\textbf{Learning from a dog.} 
We test our hypothesis about the information encoded in our representation learned from mimicking dogs behaviour by comparing our representation with a similar one trained for image classification on ImageNet on a third task. We chose the task of  walkable surface estimation and scene classification.

\textit{1) Walkable surface estimation.}
The goal for this task is to label pixels that correspond to walkable regions in an image (e.g., floor, rug, and carpet regions). We use the dataset provided by \cite{mottaghi16b}. In our dataset, we have some sequences of dog walking in indoor and outdoor scenes. There are various types of obstacles (e.g., furniture, people, or walls) in the scenes that the dog avoids. We conjecture that the learned representation for our network should provide strong cues for estimating walkable surfaces. 

The definition of walkable surfaces for humans and dogs is not the same. As an example, the area under the tables are labeled as non-walkable for humans and walkable for dogs. However, since our dog is large-size dog, the definition of walkability is roughly the same for humans and the dog.

We trained ResNet-18 on ImageNet and then finetuned it on the walkable surface dataset as our baseline. We performed the same procedure for using our features (trained for the \emph{acting} task). For finetuning both models, we just update the weights for the last convolutional layer (green block in Figure~\ref{fig:walkable}). 

Table~\ref{tab:walkable} shows the results. Our features provide a significant improvement, $3\%$, over the ImageNet features. We use IOU as the evaluation metric. This indicates that our features have some information orthogonal to the ImageNet features.

\textit{2) Scene classification.}
We perform an additional scene recognition experiment using SUN 397 dataset \cite{sun397}. We used the same 5-instance training protocol used by \cite{agrawal15}. The representation learned by us obtains the accuracy of 4.48 (as a point of reference \cite{jayaraman15} achieves 1.58 and \cite{agrawal15} achieves 0.5-6.4 from their representations, and the chance is 0.251). This is interesting since our dataset does not include many of the scene types (gas station, store, etc).

\begin{table}[t]
\centering
\begin{tabular}{|l | c | c | c | c | c | c |}
 \hline
 Model & Pre-training task &    IOU \\
 \hline
 ResNet-18 & ImageNet Classification & $42.88$\\
 \hline
 ResNet-18 & Acting like a dog & $\mathbf{45.60}$\\
 \hline
\end{tabular}
\vspace{0.1cm}
\caption{\textbf{Walkable surface estimation.} We compare the result of the network that is trained on ImageNet with the network that is trained for our \emph{acting} task. The evaluation metric is IOU.}
\vspace{-0.4cm}
\label{tab:walkable}
\end{table}

%% file: conclusion.tex
\section{Conclusion}

We study the task of directly modeling a visually intelligent agent. Our model learns from ego-centric video and movement information to act and plan like a dog would in the same situation. We see some success both in our quantitative and qualitative results. Our experiments show that our models can make predictions about future movements of a dog and can plan movements similar to the dog. 

This is a first step towards end-to-end modelling of intelligent agents. This approach does not need manually labeled data or detailed semantic information about the task or goals of the agent. We can use this model on a wide variety of agents and scenarios and learn useful information despite the lack of semantic labels.


For this work, we limit ourselves to only considering visual data. However, intelligent agents use a variety of input modalities when interacting with the world including sound, touch, smell, etc. We are interested in expanding our models to encompass more input modalities in a combined, end-to-end model. We also limit our work to modelling a single, specific dog. It would be interesting to collect data from multiple dogs and evaluate generalization across dogs. We hope this work paves the way towards better understanding of visual intelligence and of the other intelligent beings that inhabit our world.
